\documentclass[letterpaper]{article} 
\usepackage{aaai23}  
\usepackage{times}  
\usepackage{helvet}  
\usepackage{courier}  
\usepackage[hyphens]{url}  
\usepackage{graphicx} 
\usepackage{natbib}  
\usepackage{caption} 
\usepackage{algorithm}
\usepackage{algorithmic}
\usepackage{multirow}
\usepackage{booktabs}
\usepackage{amssymb}
\usepackage{amsmath}
\usepackage{newfloat}
\usepackage{listings}
\DeclareCaptionStyle{ruled}{labelfont=normalfont,labelsep=colon,strut=off} 
\floatstyle{ruled}
\newfloat{listing}{tb}{lst}{}
\floatname{listing}{Listing}
\title{NeAF: Learning Neural Angle Fields for Point Normal Estimation}
\author{
    Shujuan Li\textsuperscript{\rm 1}\equalcontrib, 
    Junsheng Zhou\textsuperscript{\rm 1}\equalcontrib, 
    Baorui Ma\textsuperscript{\rm 1},
    Yu-Shen Liu\textsuperscript{\rm 1}\thanks{Corresponding author: Yu-Shen Liu}, 
    Zhizhong Han\textsuperscript{\rm 2}
}

\affiliations{
    \textsuperscript{\rm 1}School of Software, BNRist, Tsinghua University, Beijing, China\\
    \textsuperscript{\rm 2}Department of Computer Science, Wayne State University, Detroit, USA\\
    {lisj22, zhoujs21, mbr18} @mails.tsinghua.edu.cn,
    liuyushen@tsinghua.edu.cn, h312h@wayne.edu
}

\begin{document}
\frenchspacing
\maketitle

\begin{abstract}
\begin{quote}

Normal estimation for unstructured point clouds is an important task in 3D computer vision. Current methods achieve encouraging results by mapping local patches to normal vectors or learning local surface fitting using neural networks. However, these methods are not generalized well to unseen scenarios and are sensitive to parameter settings.
To resolve these issues, we propose an implicit function to learn an angle field around the normal of each point in the spherical coordinate system, which is dubbed as Neural Angle Fields (NeAF). Instead of directly predicting the normal of an input point, we predict the angle offset between the ground truth normal and a randomly sampled query normal. This strategy pushes the network to observe more diverse samples, which leads to higher prediction accuracy in a more robust manner. To predict normals from the learned angle fields at inference time, we randomly sample query vectors in a unit spherical space and take the vectors with minimal angle values as the predicted normals. To further leverage the prior learned by NeAF, we propose to refine the predicted normal vectors by minimizing the angle offsets. The experimental results with synthetic data and real scans show significant improvements over the state-of-the-art under widely used benchmarks. Project page: \url{https://lisj575.github.io/NeAF/}.

\end{quote}
\end{abstract}

\section{Introduction}

Estimating normals for unstructured point clouds is a fundamental task in 3D computer vision. The estimated normal vectors can be leveraged in various downstream applications, such as surface reconstruction \cite{MichaelKazhdan2006PoissonSR}, registration \cite{pomerleau2015review}, and semantic segmentation \cite{EleonoraGrilli2017AReviewOP}.  
Recently, the learning-based methods have achieved promising results by casting the normal estimation task into a regression problem. They resolved this problem by directly learning a  mapping from local patches to normal vectors using neural networks. However, these methods fail to observe adequate samples for accurate normal estimation. This leads to poor generalization ability on unseen scenarios such as large-scale scenes.

As a remedy, state-of-the-art methods introduce to predict the normal of a point by fitting a local geometric surface. They learned point-wise weights for neighboring points to fit the surface, and then predict the normal vector from the fitted surface. However, the geometric surface settings (e.g. the constant order of the polynomial surface) are predefined during the training process, which leads to poor fitting results due to the various complexity of the local point patches.
For example, when the geometric surface is more complex than the underlying point patches, it results in overfitting problems. Otherwise, the underfitting leads to over-smoothing details. As a result, the performance of surface fitting methods is dramatically limited in cases with complex topology and geometry.

\begin{figure*}[t]
\centering
\includegraphics*[width=1.0\linewidth]{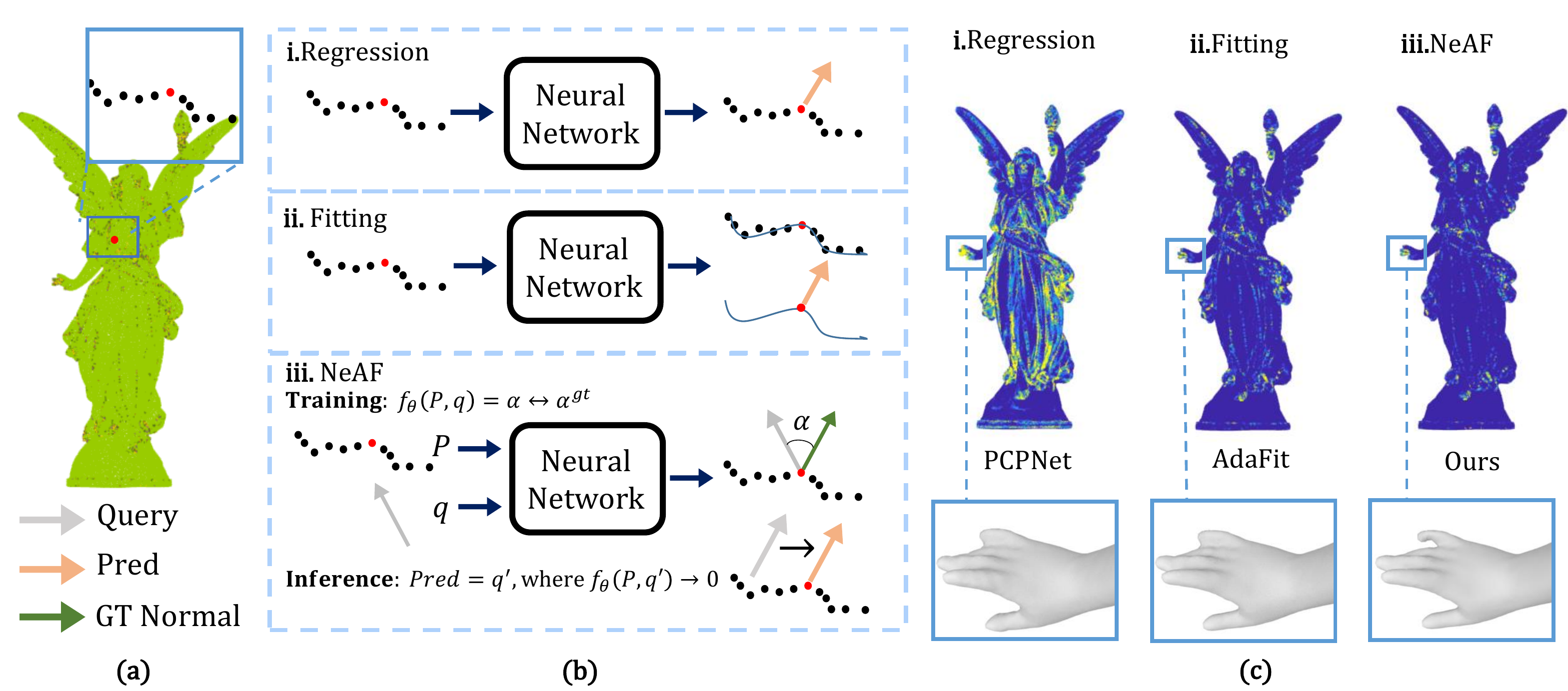}
\caption{Comparisons between previous learning-based normal estimation approaches and our NeAF. (a) Given a point on the point cloud, we sample its K nearest neighbors as a local patch and output the normal for the point. (b) Existing learning-based normal estimation methods can be roughly divided into (i) regression-based and (ii) fitting-based methods. (i) Regression-based methods directly map the local patch to a 3D vector as the normal. (ii) Surface fitting-based methods estimate the weight for each point in the local patch and compute the normal of the surface fitted with the weighted points. (iii) Our NeAF learns the angle offsets between the random query vectors and the ground truth normal, and outputs the query vector with an angle offset zero as the normal vector. (c) The colors of the shapes indicate normal RMSE errors. The estimated normals can be used for surface reconstruction and significantly affect the quality of the reconstructed surface.}
\label{fig:methods_cpmparison}
\end{figure*}

To resolve these issues, we propose an implicit method to learn the Neural Angle Field (NeAF) of local patches in the spherical coordinate system. Specifically, instead of explicitly predicting the normal from the input point clouds, we design a neural network to implicitly predict the angle offset between the ground truth normal and a randomly sampled query vector. The network observes more diverse samples by learning to model the relationship between vectors from various directions and the target normal, leading to an angle field around the target normal.
With a well-learned angle field, the network will have the ability to predict accurate and robust normal vectors. 

At inference time, to predict normals from the learned angle fields, we randomly sample query vectors in a unit spherical space and take the vectors with minimal angle offsets as the predicted normals. To further leverage the prior learned by NeAF, we propose a novel learning scheme for refining the predicted normal vectors. Unlike existing learning-based methods that can only predict the normal via one forward pass, our proposed method can further optimize the predicted normal, which leads to more accurate normal estimation. 

Our contributions are summarized as follows.
\begin{itemize}
\item We propose Neural Angle Field (NeAF) for point normal estimation. Unlike previous methods, our implicit function learns an angle field for each point, which implicitly predicts the angle offset of query vectors.
\item Our method can conduct further optimization at inference time to refine the predicted normals by minimizing the angle offsets for more accurate normal estimation.
\item We achieve state-of-the-art results in normal estimation for synthetic data and real scans on widely used benchmarks.

\end{itemize}

\section{Related Work}

\subsection{Normal Estimation}

Traditionally, principal component analysis (PCA) \cite{HoppeHugues1992SurfaceRF} is a simple and efficient method for normal estimation, which is based on constructing a tangent plane from a fixed-scale local neighborhood and analyzes its normal vector.  Variants such as truncated Taylor expansion (Jets) \cite{FrdricCazals2003EstimatingDQ}, moving least squares (MLS)  \cite{DavidLevin1998TheAP} and multi-scale kernel methods \cite{SamirAroudj2017VisibilityconsistentTS} were proposed to fit more complex local surfaces. However, these methods tend to choose large scales to ensure the robustness of normal estimation. To preserve more details, the approaches \cite{NinaAmenta1998SurfaceRB,QuentinMrigot2011VoronoiBasedCA} were proposed based on analyzing Voronoi cells. In practice, these methods have to tune their parameters to work with different cases, which dramatically limits their application scenarios.

Recently, with the development of deep learning, learning-based methods were proposed for normal estimation. Boulch and Marlet  \shortcite{AlexandreBoulch2016DeepLF} proposed to transform the local patches of point clouds into accumulators in a 2D Hough space and estimate normals from the resulting approximate planes. However, the transformation from 3D to 2D resulted in the loss of the structural information contained in the surface. To preserve a complete local shape, Guerrero et al. \shortcite{PaulGuerrero2017PCPNETLL} proposed the PCPNet with a multi-scale point cloud normal estimation architecture using PointNet \cite{CharlesRQi2016PointNetDL}. Based on this method, Zhou et al. \shortcite{JunZhou2019NormalEF} proposed a new scale selection strategy and extra constraints on the feature. In addition, Zhou et al. \shortcite{9693131} proposed to introduce additional feature representations to refine the input initial normal. Li et al. \shortcite{li2022hsurf} used the networks to learn hyper surfaces and obtained an improved performance.

Recent studies have achieved encouraging performance on normal estimation by predicting point-wise weights to select neighboring points softly. Lenssen et al. \shortcite{JanEricLenssen2019DeepIS} proposed to use a graph neural network to iteratively generate point-wise weights, and then estimate the normal by fitting a moving least squares plane. Ben-Shabat et al. \shortcite{ben2020deepfit} used the n-order Jet of weighted points to fit local surfaces. Cao et al. \shortcite{JunjieCao2022LatentTS} used a differentiable random sample consensus module to predict normals from a latent tangent plane representation constructed from the neighboring points. To learn better point-wise weights, Zhang et al.  \shortcite{JieZhang2022GeometryGD} proposed to add direct geometric weight guidance which is constructed by distances between points and the tangent plane. Zhu \shortcite{RunsongZhu2021AdaFitRL} proposed to predict an offset to adjust the distribution of clouds and designed a cascaded scale aggregation (CSA) layer adapting to the scale of the local neighborhood, which achieved state-of-the-art results. Li et al. \shortcite{li2022graphfit} proposed to learn graph convolutional feature representation. Although improvements have been made by predicting point-wise weights and fitting surfaces for normal estimation, these methods still suffer from several inherent problems such as the difficulty in determining the form of the fitted surface and the sensitivity to outliers.

\subsection{Neural Implicit Representation in 3D Reconstruction}
Recently, neural implicit representations have gained popularity and are widely used for 3D reconstruction \cite{Zhou2022CAP-UDF, On-SurfacePriors, LPI, PredictableContextPrior, Li2022DCCDIF}. These representations were learned by neural networks to map input 3D coordinates to occupancy probabilities \cite{chen2019learning, mescheder2019occupancy} or distance values \cite{MateuszMichalkiewicz2019ImplicitSR,JeongJoonPark2019DeepSDFLC}, which implicitly represent continuous surfaces of shapes and handle complex shape topology. In addition, Neural Radiance Fields (NeRFs) \cite{mildenhall2020nerf} implicitly encode the geometry and color information of the scene in the network.

Inspired by neural implicit representations, we propose to utilize the network to implicitly represent the normals of points. Unlike existing learning-based normal estimation methods that focus only on the relationship between local neighborhoods and normal vectors, we use the network to observe the difference between random vectors and the ground truth normal. This provides the network with more prior knowledge by learning a neural angle field, which helps to predict more accurate and robust normals.

\section{Method}
In this section, we present NeAF, an implicit normal estimation method for point clouds.

\begin{figure}[t]
\centering
\includegraphics*[width=1.0\linewidth]{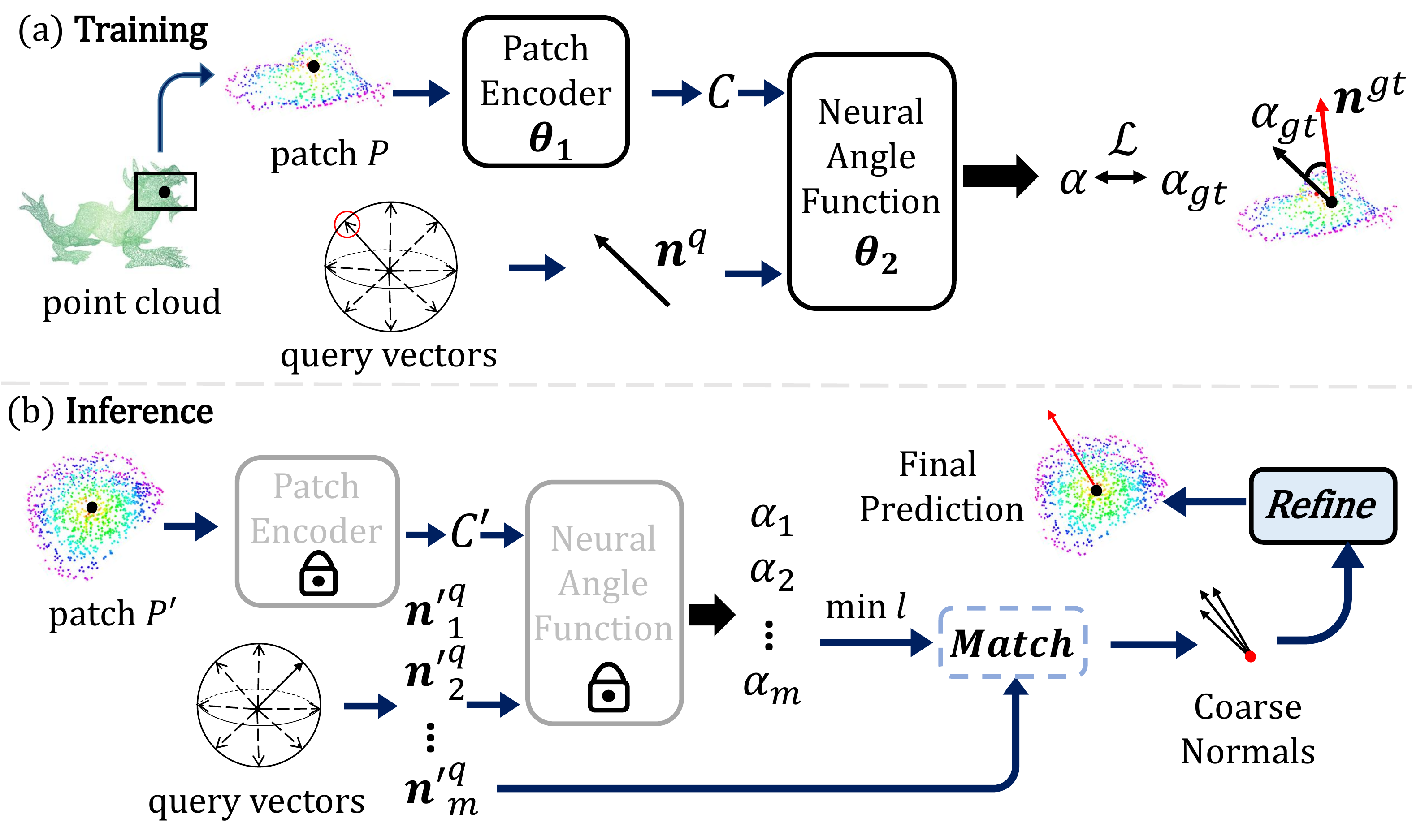}
\caption{Overview of our NeAF. (a) During training, we randomly sample query vectors in a unit spherical space, together with the local patch as the input to the network, and predict the angle offset $\alpha$ between the query vector $\boldsymbol{n}^q$ and the ground truth normal $\boldsymbol{n}^{gt}$. (b) At inference time, we sample $m$ query vectors at different directions, and then select the query vectors with the minimum $l$ angle offsets as the coarse normals. \emph{Match} means to find the query vectors with the same indices as the minimum $l$ angle offsets. \emph{Refine} means coarse normal refinement in a subsequent section.}
\label{fig:overview}
\end{figure}

\subsection{Implicit Angle Functions}
We first define the \emph{implicit angle function} of a given normal vector. The implicit angle function takes an arbitrary query vector in a unit spherical space and a condition $\mathcal{C}$ as input and predicts the angle offset between the query vector and the ground truth vector. The corresponding function is defined as follows,
\begin{equation}
\label{eq:definition of angle function}
f: \mathcal{C} \times \mathbb{R}^3   \to \mathbb{R}.
\end{equation}

Our key idea is to estimate the normal of one point using the implicit angle function that is learned by a neural network. Observing that the normal vector is a local property of a surface, we leverage the local neighborhood of the center point as a condition $\mathcal{C}$ indicating the center point, together with the query vector as the input to feed the network, and the output is the angle offset between the ground truth and the query vector. Since we focus on unoriented normal estimation, the angle offset is either the difference from the positive direction or the negative direction of the ground truth, and its value range is $[0, \frac{\pi}{2}]$. The implicit angle function of point $p_i$ can be represented as:
\begin{equation}
\label{eq:angle function for network}
f_\theta(P_i, \boldsymbol{n}^q) = \alpha, i \in [1,N],
\end{equation}
where $f_\theta$ denotes the network parameterized by $\theta$,  $P_i \in \mathbb{R}^{k \times 3}$ is the condition of point $p_i$, it is the set of $k$ nearest neighbors for the center point $p_i$, $N$ is the number of points on the point cloud, $\boldsymbol{n}^q \in \mathbb{R}^3$ is the query vector, and $\alpha  \in [0, \frac{\pi}{2}]$ is the angle offset of $\boldsymbol{n}^q$ relative to the target normal. 

With this definition, the target normal of $p_i$ is implicitly represented as the zero level set of the implicit angle function, which is denoted as $\boldsymbol{n}^{t}$, i.e.  $f_\theta(P_i, \boldsymbol{n}^{t}) = 0$.

\subsection{Training for Implicit Angle Functions}
To learn the parameters $\theta$ of the neural network $f_\theta$, we first randomly sample query vectors $\boldsymbol{n}^q_j, j=1,...M$ on the unit sphere. For a given point $p_i, i=1,...N$, we use the K-Nearest-Neighbors algorithm to sample the nearest neighbors $P_i$ as the local patch centered at $p_i$. Next, the network takes the local patch $P_i$ and query vector $\boldsymbol{n}^q_j$ as inputs and predicts the angle offsets between the ground truth normal $\boldsymbol{n}^{gt}_{i}$ and $\boldsymbol{n}^q_j$. We minimize the difference between the predicted angle offset $f_\theta(P_i, \boldsymbol{n}^q_j)$ and ground truth angle offset $\alpha^{gt}_{ij}$ by the following $L1$ loss,
\begin{equation}
\label{eq:loss}
\mathcal{L}= \frac{1}{NM}\sum_{i \in [1,N]}{\sum_{j\in [1,M]}}|f_\theta(P_i, \boldsymbol{n}^q_j) - \alpha^{gt}_{ij}|,
\end{equation}
and the $\alpha^{gt}_{ij}$ is computed by,
\begin{equation}
\label{eq:compute gt}
\alpha^{gt}_{ij} = arcsin(||\boldsymbol{n}^{gt}_{i}\times \boldsymbol{n}^q_j||),
\end{equation}
where $arcsin(\cdot)$ represents the arcsine function, $||\cdot||$ is the L2-norm, and $\times$ is cross product.

\begin{figure}[t]
\centering
\includegraphics*[width=1.0\linewidth]{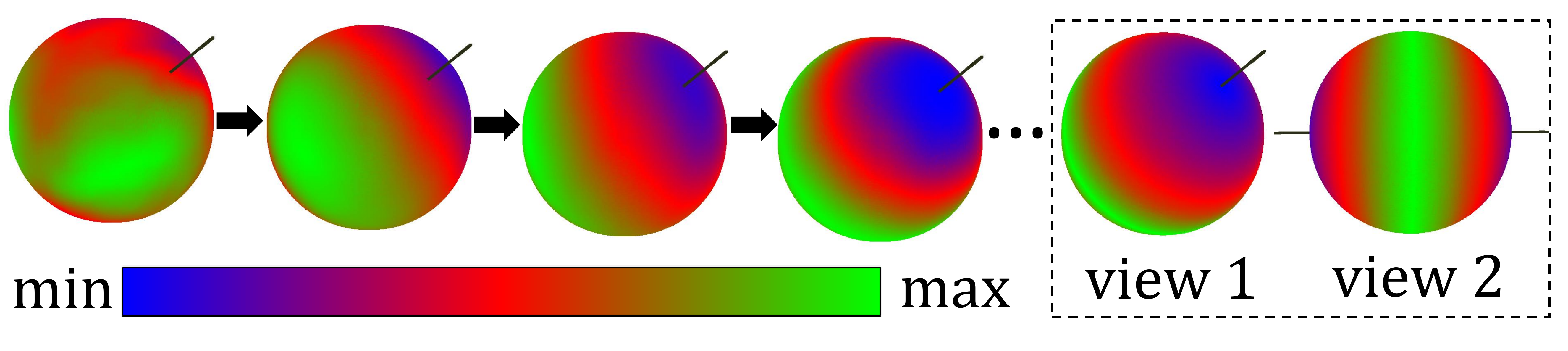}
\caption{The visualization of learning an implicit angle field. Each point on the unit sphere represents a query vector, and the color map indicates their angle offsets to the ground truth normal represented as the black line. After convergence, an angle field centered on the ground truth normal is formed.}
\label{fig:field visualization}
\end{figure}

After convergence, the network learns an implicit angle field for each point. In Figure \ref{fig:field visualization}, we visualize the optimizing process of the angle field around a single point, where the well-learned angle field can predict correct angle offsets on the whole spherical surface. 

The angle field implicitly represents the target vector as the zero level-set of the learned angle field. At inference time, to predict the target vector from the learned field, we introduce a gradient descent based optimization to explore the prior learned by the angle field. An intuitive implementation is to directly optimize some randomly sampled vectors which cover the whole spherical space until their angle offsets approach zero, but it increases the computational burden to optimize a large number of vectors during inference. To resolve this issue, we propose to first predict a few coarse normals and then optimize these predicted coarse normals by a gradient decent based optimization algorithm. 

\subsection{Inference of Angle Fields}

\subsubsection{Coarse Normal Prediction.}
To predict coarse normals at inference time, we discretize the continuous unit spherical space into sections, which are the different query directions. Then a certain density of points are sampled from the discrete sections to form a query vector set $S=\{{{\boldsymbol{n}}^q_k}' | k =1, 2, ... , m\}$ to cover $m$ different directions. 
If the query vectors in $S$ are infinitely dense to cover the full spherical space, there is a vector $\boldsymbol{n}^o \in S$ that satisfies $f_{\theta}(P', \boldsymbol{n}^o) = 0$, then $\boldsymbol{n}^o$ can be viewed as the approximated direction of ${\boldsymbol{n}^{gt}}'$ .

However, it is impossible to cover the full spherical space during testing since the discrete query vectors approximate the continuous spherical space and it is also time-consuming to infer highly dense query vectors. Therefore, we propose to predict a few coarse normals from the learned angle fields first.

In practice, given a test point $p_i'$, the trained network $f_{\theta}$ takes the local neighborhood $P_i'$ around $p_i'$ and the query vectors in $S$ as input, and predicts the angle offsets $ A = \{\alpha_{ik} | k = 1,2,...,m\}$, that is $f_{\theta}(P'_i, {{\boldsymbol{n}}^q_k}') = \alpha_{ik}$. We select the query vectors $\{ \boldsymbol{n}^c_s|s=1,...,l\}$ from $S$ with the minimum $l$ angle offsets as the predicted coarse normals. The coarse normals will be further refined in the following.

\begin{figure}[t]
\centering
\includegraphics*[width=1.0\linewidth]{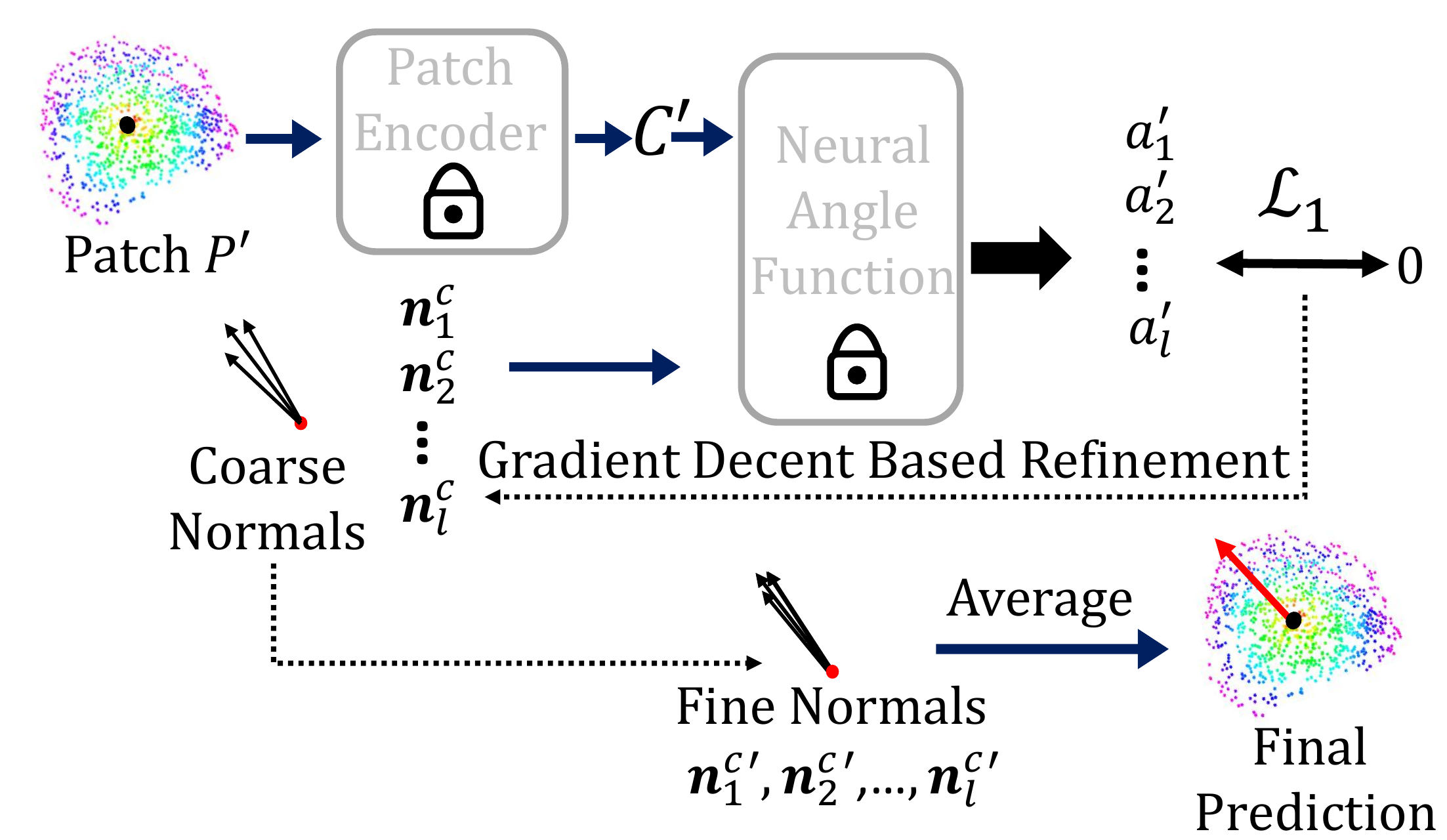}
\caption{Coarse normal refinement. For the coarse normal $\boldsymbol{n}^c_{s},s=1...l$, we use the trained network $f_{\theta}$ with fixed parameters to estimate the angle offset $\alpha_s'$, then we update the vector $\boldsymbol{n}^c_s$ to make $\alpha_s'$ closer to zero. When the further optimization is done, we average all the coarse normals to output the final predicted normal.}
\label{fig:self-optimization}
\end{figure}

\begin{table*}[pt]
    \small    
    \centering
    \begin{tabular}{l|cc|cccc|l}  
    \toprule 
    \multirow{2}*{Methods} &\multicolumn{2}{c}{Density}   &\multicolumn{4}{c|}{Noise} & \multirow{2}{*}{average} \cr  \cmidrule{2-7}
    & Stripes &Gradients & No noise & Low Noise& Med Noise& High Noise\cr

    \midrule 
    Jets \cite{FrdricCazals2003EstimatingDQ}
    &13.39 &13.13& 12.23 &12.84 &18.33 &27.68 & 16.29\cr
    PCA \cite{HoppeHugues1992SurfaceRF}
    &13.66 &12.81&12.29 &12.87 &18.38 &27.5 & 16.25\cr
    HoughCNN \cite{AlexandreBoulch2016DeepLF}
    &12.47&11.02&10.23&11.62&22.66&33.39&16.90\cr
    PCPNet \cite{PaulGuerrero2017PCPNETLL}
    &11.74 &13.42&9.66 &11.46 &18.26 &22.8 & 14.56\cr
    Nesti-Net \cite{YizhakBenShabat2018NestiNetNE}
    &8.47&9.00&6.99&10.11&17.63&22.28&12.41\cr
    IterNet \cite{JanEricLenssen2019DeepIS}
    &7.73&7.51&6.72&9.95&17.18&21.96&11.84\cr
    DeepFit \cite{ben2020deepfit}
    &7.31&7.92&6.51&9.21&16.72&23.12&11.8\cr
    Refine-Net \cite{9693131}
    &6.61&7.02&6.27&9.18&16.59&22.57&11.37\cr
    AdaFit \cite{RunsongZhu2021AdaFitRL}
    &6.04&5.90&5.19&\textbf{9.05}&16.44&21.94&10.76\cr
    NeAF (Ours)
    &\textbf{4.89}&\textbf{4.88}& \textbf{4.20}&9.25&\textbf{16.35}&\textbf{21.74}&\textbf{10.22}\cr 
    \bottomrule
    \end{tabular}
    \caption{Comparison of the angle RMSE with the state-of-the-art methods on PCPNet dataset.}
    \label{tab:RMSE on PCPNet} 
\end{table*}

\subsubsection{Coarse Normal Refinement.}
The analysis of normal prediction by the learned angle fields shows that the random sampling of discrete spherical points cannot perfectly cover the entire continuous spherical space, thus the forward pass prediction cannot make full use of the learned angle fields.
To further improve the accuracy of the predicted normals at inference time, we propose a novel learning scheme for refining the predicted normal vectors by a gradient decent based optimization algorithm. Unlike existing learning-based methods that can only predict normals via a single forward pass during inference, our proposed method can fully leverage the prior learned by NeAF to adjust the predicted normals, and can further optimize the predicted normal at inference time, thus leading to a more accurate estimation.

Specifically, the proposed NeAF represents the relationship between the target normal and an arbitrary vector in the neural network by measuring the angle offset. Thus we can adopt further refinement given the predicted coarse normals by minimizing the angle offsets of these normals.

At inference time, we utilize the network $f_{\theta}$ with fixed parameters to output an angle offset $\alpha'_{is} $ for the coarse normal $\boldsymbol{n}^c_s$, and then update $\boldsymbol{n}^c_s$ to make $\alpha'_{is}$ as close as possible to zero following the formula,
\begin{equation}
\label{eq:optimization}
    {\boldsymbol{n}^c_s}' = \arg \min \limits_{\boldsymbol{n}^c_s} \mathcal{L}_1(f_{\theta}(P'_i, \boldsymbol{n}^c_s)-0),
\end{equation}
where $\mathcal{L}_1$ is the $L1$ loss, and $P'_i$ is the local patch for the testing point $p_i'$.

After coarse normal refinement, $\boldsymbol{n}^c_s$ is optimized to approximate the real zero level-set of the learned angle field. 

To ensure the robustness of the results, we propose to average these refined normals ${{\boldsymbol{n}}^c_s}', s\in[1,l]$ as the final estimation $\boldsymbol{n}_{pred}$ by the following formula,
\begin{equation}
\label{eq:average coarse normals}
    \boldsymbol{n}_{pred} = \frac{1}{l}\sum_{s \in [1,l]}{{\boldsymbol{n}}^c_s}'.
\end{equation}

Since the correct normal vector direction can be easily estimated by several methods \cite{mullen2010signing,  wu2015deep, huang2019learning, metzer2021orienting}, we mainly focus on \emph{unoriented normal estimation}. However, the mean of the refined normals described above may lead to a large error due to the uncertain signs. For example, both the vector with an actual angle offset of $\hat{\alpha}$ and the one with an angle offset of $(\pi-\hat{\alpha})$ are with the same angle offset $\hat{\alpha}$, but the mean of the two normals can be far from the ground truth normal by an offset of $\pi/2$. To solve this problem, we normalize the signs of these vectors before averaging them together. We choose one of the refined vectors ${{\boldsymbol{n}}^c_r}'$ as the reference  and re-direct the other vectors ${{\boldsymbol{n}}^c_s}', s\in[1,l]$, according to the dot product between ${{\boldsymbol{n}}^c_r}'$ and ${{\boldsymbol{n}}^c_s}'$ below
\begin{equation}
\label{eq: sign}
    {{\boldsymbol{n}}^c_s}' = sign ({{{\boldsymbol{n}}^c_r}'}^T {{\boldsymbol{n}}^c_s}'){{\boldsymbol{n}}^c_s}',
\end{equation}
where $sign$ represents the sign function with an output in $\{-1, 1\}$.

\subsection{Implementation Details}
For generating training data, we adopt the method proposed by Muller \shortcite{muller1959note} to randomly and uniformly sample $M=5000$ query vectors in the unit spherical space for training, and the ground truth angle offsets are calculated by Eq. (\ref{eq:compute gt}). As for inference data, we use the same method to sample $m=10000$ query vectors for extracting coarse normals. During training, we randomly select 400 query vectors from the training set as a batch to train the network. At inference time, we select to extract $l=10$ coarse normals and optimize them simultaneously in 5 epochs for coarse normal refinement. We use the same strategy as AdaFit \cite{RunsongZhu2021AdaFitRL} for preprocessing local patches, including centering and normalizing.

We employ an encoder similar to AdaFit to learn features of local patches, which is mainly based on the PointNet  \cite{CharlesRQi2016PointNetDL} architecture, and we retain the CSA layer. The local patch size and parameter settings of our networks are also the same as AdaFit. Besides, we adopt a neural network similar to DeepSDF \cite{JeongJoonPark2019DeepSDFLC} to decode the angle offsets, which is composed of 8 fully connected layers with a residual connection.

\section{Experimental Results}

\subsection{Evaluation on synthetic dataset}

\subsubsection{Dataset and metrics.}
For the experiments on synthetic shapes, we adopt the PCPNet dataset provided by Guerrero et al. \shortcite{PaulGuerrero2017PCPNETLL}. We use the same train/test settings and data augmentation strategies. PCPNet samples 100k points on the mesh of each shape to obtain a point cloud. The training set contains 8 shapes, and each shape includes a noise-free point cloud and three point clouds containing Gaussian noise with a standard deviation of 0.125\% (Low), 0.65\% (Med) and 1.2\% (High) of the length of the bounding box diagonal of the shape. In addition to the three noise variants, two additional point clouds with varying densities (Stripes and Gradients) are added to the test set. For training, we use the Adam optimizer with an initial learning rate of $1\times10^{-3}$, and adopt a cosine learning rate decay strategy with warm-up. The model is trained on 2 GTX 1080Ti. In coarse normal refinement, we use the Adam optimizer with an initial learning rate of 0.005. We use the angle root mean square error (RMSE) between the predicted normal and the ground truth normal as a quantification metric, and compute the final result with 5000 points subset sampled by PCPNet.

\subsubsection{Comparisons.}
We make a comparison with traditional normal estimation methods, including PCA \cite{HoppeHugues1992SurfaceRF}, Jets  \cite{FrdricCazals2003EstimatingDQ}, and learning-based methods HoughCNN \cite{AlexandreBoulch2016DeepLF}, PCPNet  \cite{PaulGuerrero2017PCPNETLL}, Nesti-Net \cite{YizhakBenShabat2018NestiNetNE}, IterNet \cite{JanEricLenssen2019DeepIS}, DeepFit \cite{ben2020deepfit}, Refine-Net \cite{9693131}, AdaFit \cite{RunsongZhu2021AdaFitRL}. A comparison of the angle RMSEs is shown in Table \ref{tab:RMSE on PCPNet}. The results show that our method significantly outperforms both the traditional methods and the state-of-the-art learning-based methods. Especially on the point clouds with density variations, the latest fitting-based method Adafit fails dramatically due to the point sparsity. This leads to a large discrepancy between the explicitly fitted surface and the underlying surface, while our method is not affected by point density and can make an accurate estimation.

\begin{figure}[t]
\centering
\includegraphics[width=1.0\linewidth]{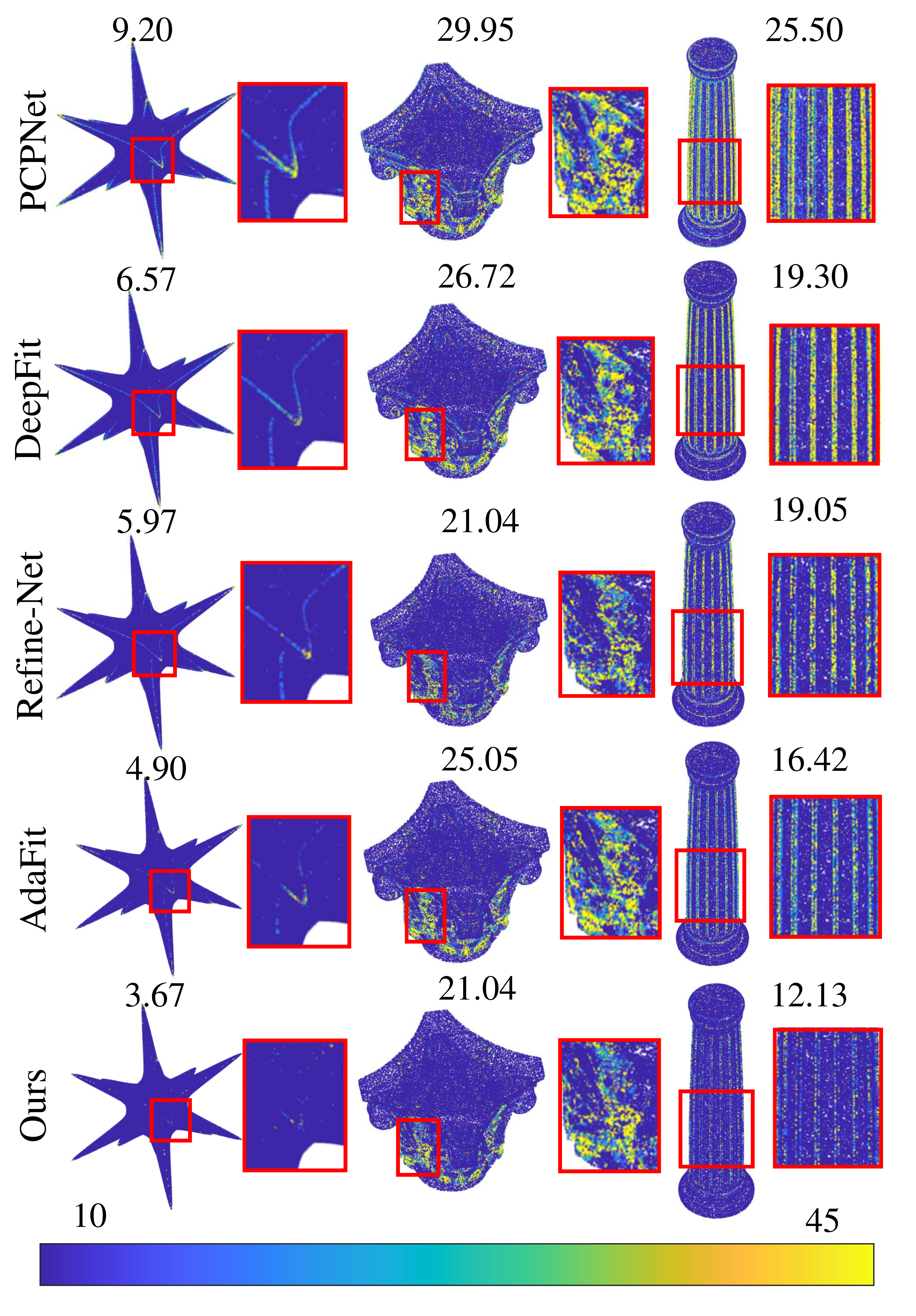}
\caption{Errors of normal estimation on the PCPNet dataset.}
\label{fig:comparision on PCPNet}
\end{figure}

We perform a visual comparison with the PCPNet \cite{PaulGuerrero2017PCPNETLL}, DeepFit  \cite{ben2020deepfit}, RefineNet \cite{9693131}, and AdaFit  \cite{RunsongZhu2021AdaFitRL} in Figure \ref{fig:comparision on PCPNet}, where the numbers represent the angle RMSE between the predicted normals and the real normals. The color of the shape indicates errors, and the closer to yellow the larger the error, the closer to the blue the smaller the error. The results show that our method outperforms others with more details, and can better handle complex regions such as sharp corners and sharp edges.

\subsection{Evaluation on real scanned dataset}
We employ the model pretrained on PCPNet dataset to report our results on the scanned dataset.

\subsubsection{SceneNN dataset.} 
\begin{figure}[!tb]
\centering
\includegraphics[width=0.9\linewidth]{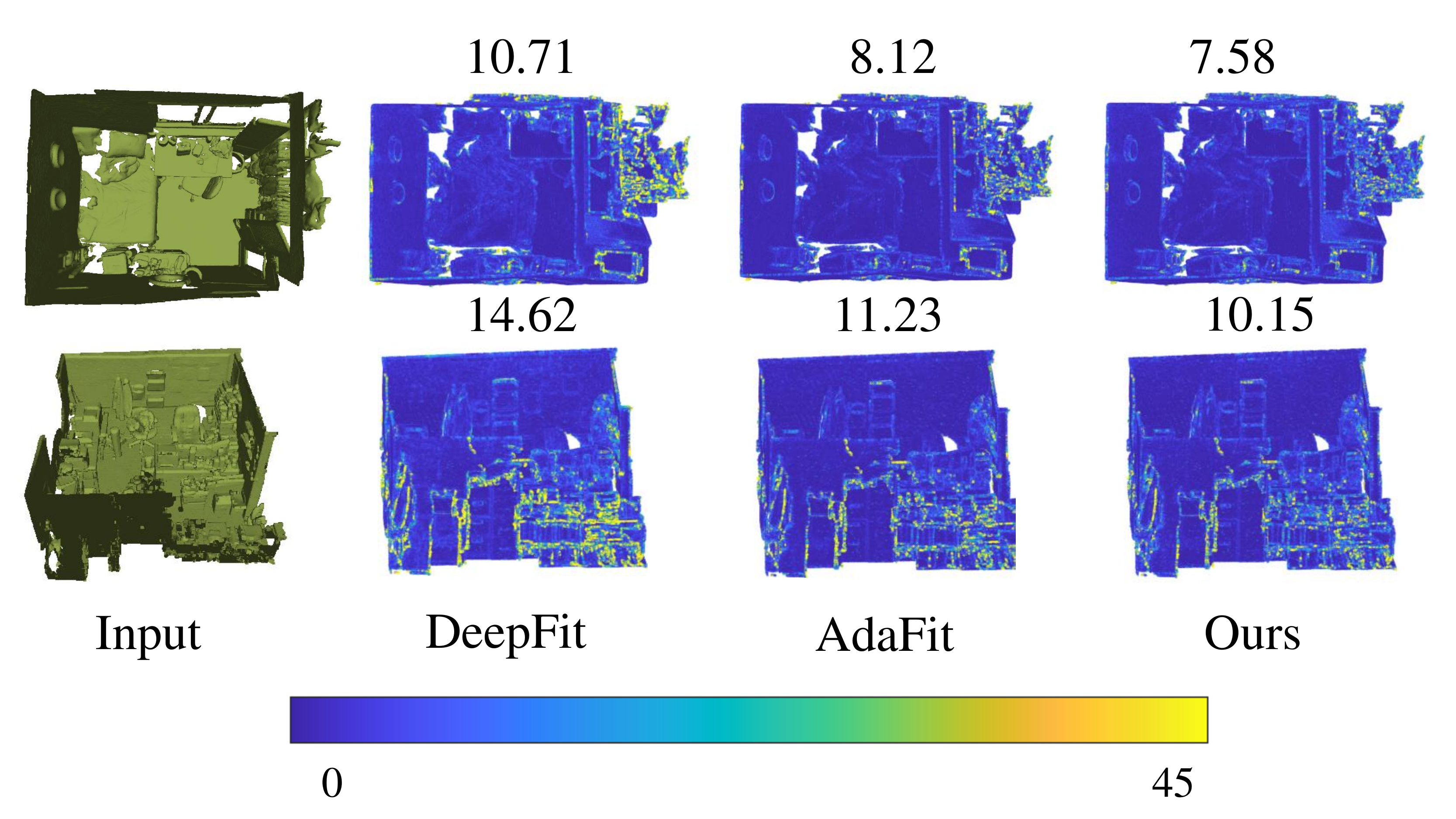}
\caption{ Error maps of estimated normals on the SceneNN dataset.}
\label{fig:comparision on SceneNN}
\end{figure}

The SceneNN \cite{BinhSonHua2016SceneNNAS} dataset provides indoor scenes in the form of reconstructed meshes. We randomly sample 1 million points on the mesh of the shape to obtain a point cloud, and calculate the normal of each point. We randomly select 40\% of all points to calculate RMSE. The results of the baseline methods are obtained by testing on the dataset using the models provided by \cite{ben2020deepfit} and \cite{RunsongZhu2021AdaFitRL} under the same experimental conditions. The angle RMSE comparison between NeAF, AdaFit \cite{RunsongZhu2021AdaFitRL} and DeepFit \cite{ben2020deepfit} is given in Table \ref{RMSE_error_indoor}, and the qualitative results are shown in Figure \ref{fig:comparision on SceneNN}. The numerical and visual comparisons show that NeAF achieves the best performance.

\begin{table}[!tb]
\centering
\begin{tabular}{cccc}
\toprule
& {NeAF} &{AdaFit}&{DeepFit} \cr
 \midrule 
RMSE & \textbf{9.50}& 10.19& 12.56\cr
\bottomrule
\end{tabular}
\caption{Normal RMSE of NeAF, AdaFit \cite{RunsongZhu2021AdaFitRL} and DeepFit \cite{ben2020deepfit} on the SceneNN dataset.}
\label{RMSE_error_indoor}
\end{table}

\begin{figure}[!tb]
\centering
\includegraphics[width=1.0\linewidth]{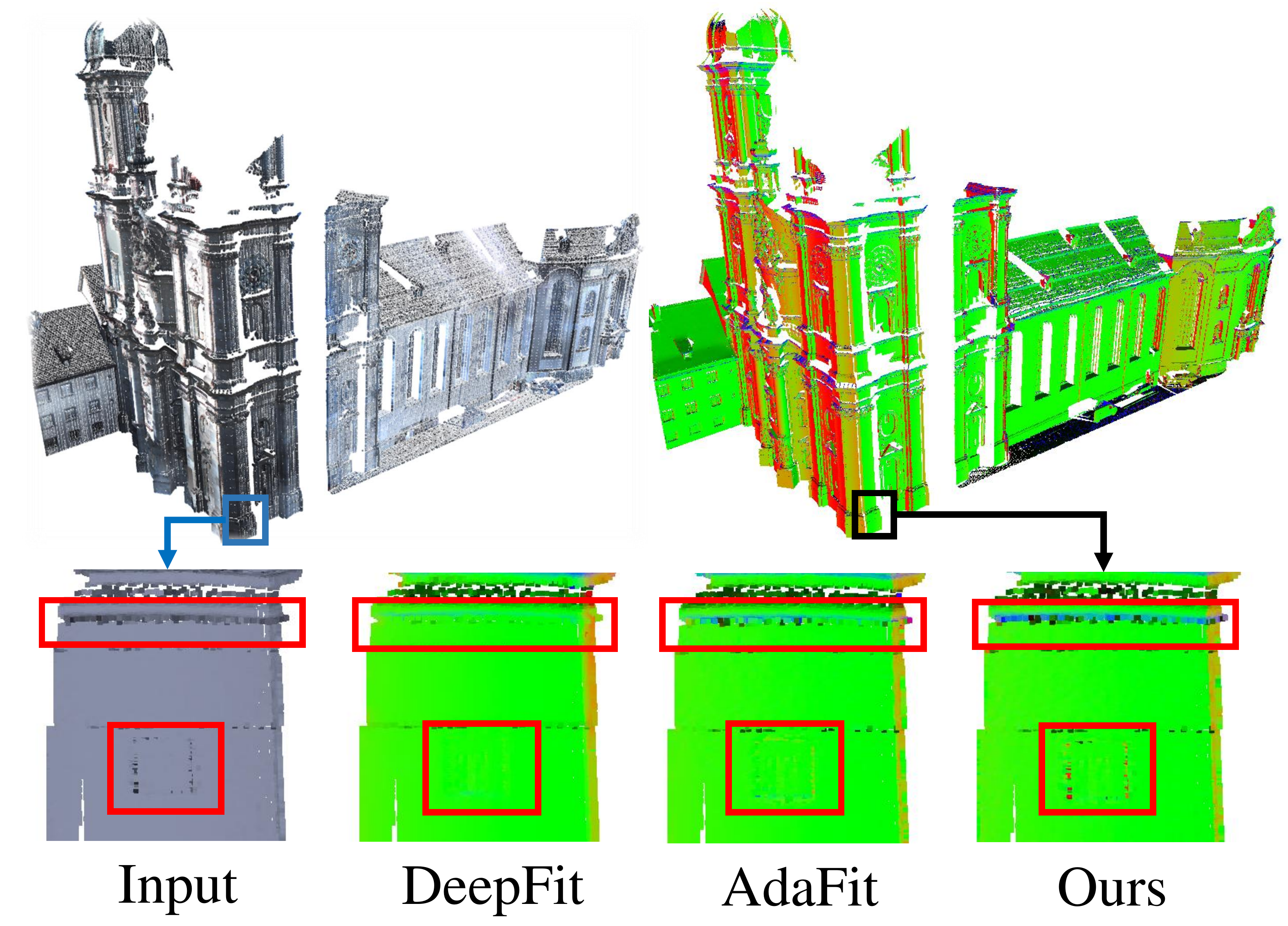}
\caption{Estimated normals on the semantic3D dataset. The point normal vectors are mapped to RGB colors.}
\label{fig:comparision on Semantic3D}
\end{figure}
\subsubsection{Semantic3D dataset.} The Semantic3D \cite{TimoHackel2017Semantic3DnetAN} dataset provides 30 non-overlapping outdoor scenes acquired with the Terrestrial Laser Scanner in the form of point clouds. This dataset does not provide reconstructed meshes, so the ground truth normals are not available, and we mainly report visual comparisons on this dataset. The comparison between NeAF and baseline methods is shown in Figure \ref{fig:comparision on Semantic3D}. The results show that the proposed method can estimate normals at sharp edges more accurately than baseline methods.

\subsection{Surface Reconstruction}
\begin{figure}[!tb]
\centering
\includegraphics[width=1.0\linewidth]{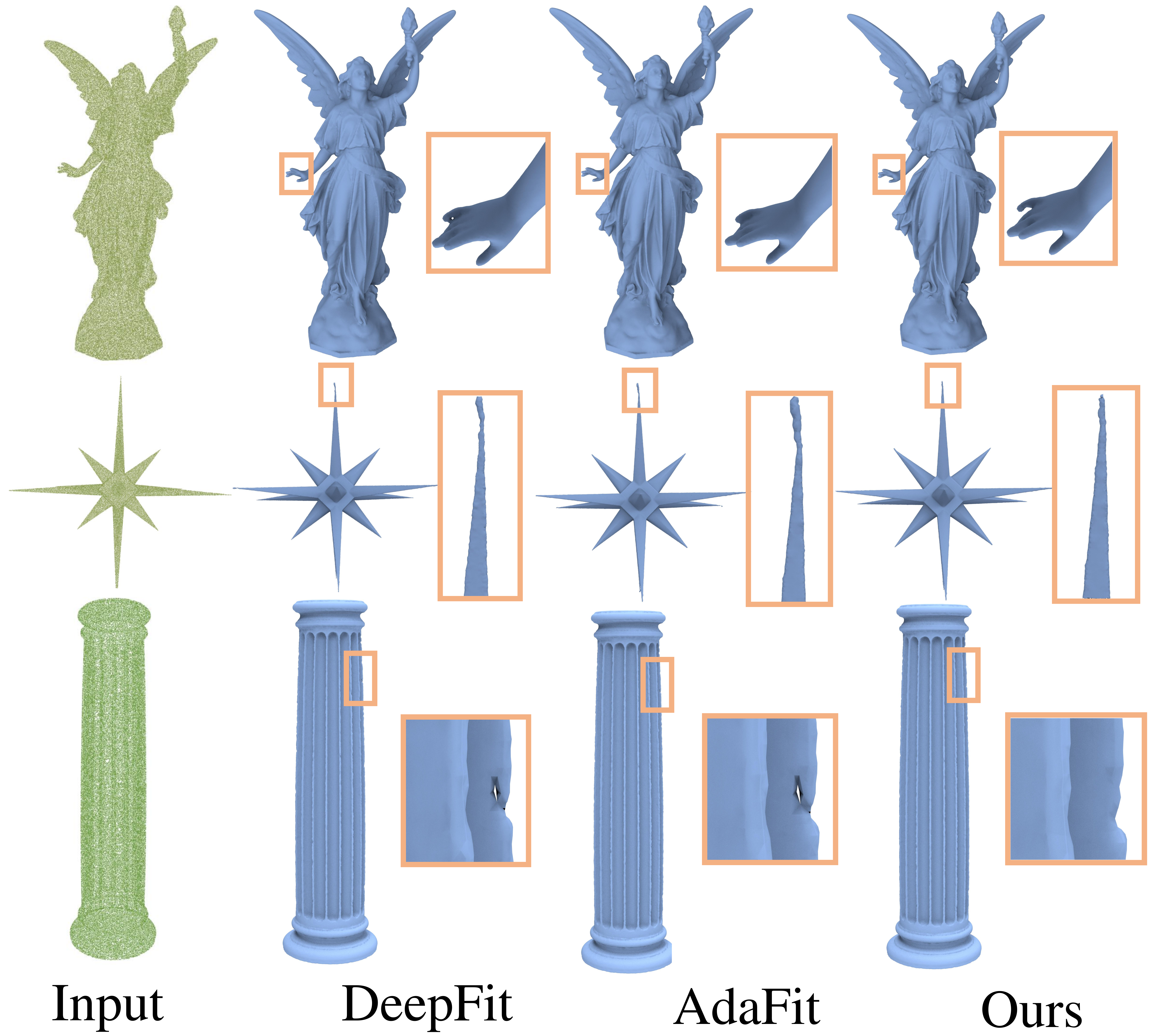}
\caption{The comparison of the Poisson surface reconstruction using the estimated normals from NeAF, DeepFit \cite{ben2020deepfit}, and AdaFit \cite{RunsongZhu2021AdaFitRL}.}
\label{fig: reconstruction}
\end{figure}

As an important property of local surfaces, normals are often used as the input for 3D surface reconstruction tasks, and accurate normal vectors play a key role in Poisson reconstruction \cite{MichaelKazhdan2006PoissonSR}. We use the estimated normals for surface reconstruction, and the comparison with the baseline methods is shown in Figure \ref{fig: reconstruction}. The results show that the normals estimated by NeAF are more accurate, which helps to reconstruct surfaces with higher accuracy.

\subsection{Ablation Studies}
\begin{table}[tb]
\centering
\begin{tabular}{l|cccc}
\toprule
 & No CNP & No CNR & Min & NeAF \cr
\midrule
Stripes & 37.23 & 5.05 & 4.94 & \textbf{4.89} \cr
Gradients & 37.03 & 5.04 & 4.93 & \textbf{4.88} \cr
No Noise & 37.11 & 4.41 & 4.26& \textbf{4.20} \cr
Low Noise & 37.97 & 9.34 & 9.27 & \textbf{9.25} \cr
Med Noise & 40.10 & 16.40 & 16.36 & \textbf{16.35} \cr
High Noise & 41.92 & 21.76 & \textbf{21.73} & 21.74 \cr
\cmidrule{1-5}
Average & 38.56 & 10.33 &  10.25 & \textbf{10.22} \cr
\bottomrule
\end{tabular}
\caption{Effect of framework design.}
\label{ablation_designs} 
\end{table}

\subsubsection{Framework design.} 
We demonstrate the effectiveness of each design of our framework in Table \ref{ablation_designs}. We first skip predicting the coarse normals and instead directly refine the random vectors the same number of times. As shown by ``No CNP'' , the performance degenerates dramatically. We also predict normals without coarse normal refinement, and the result demonstrates that the coarse normal refinement can effectively improve the accuracy of the predicted normals in all settings as shown by ``No CNR''. We replace averaging coarse normals with selecting the coarse normals  with the minimum angle offsets after refinement as shown by ``Min'' and find a drop in performance.

\begin{table}[tb]
\centering
\begin{tabular}{l|cccc}
\toprule
$M$ & 2.5k & 5k & 7.5k & 10k \cr
\midrule
Stripes & 4.94 & \textbf{4.89} & 4.95 & 4.96 \cr
Gradients & 4.98 & \textbf{4.88} & 4.98 & 5.03 \cr
No Noise & 4.29 & \textbf{4.20} & 4.24 & 4.37 \cr
Low Noise & 9.17 & 9.25 & 9.19 & \textbf{9.16} \cr
Med Noise & 16.41 & \textbf{16.35} & 16.49 & 16.41 \cr
High Noise & 21.83 &21.74 & 21.81 &  \textbf{21.69} \cr
\cmidrule{1-5}
Average & 10.27 & \textbf{10.22} & 10.28 & 10.27 \cr
\bottomrule
\end{tabular}
\caption{Effect of query vector number $M$.}
\label{ablation_density} 
\end{table}

\begin{table}[tb]
\centering
\begin{tabular}{l|cccc c}
\toprule
$l$ & 1 & 5 & 10 & 20 & 50 \cr
\midrule
Stripes & 4.95 & 4.90 & 4.89 & 4.89& 4.89\cr
Gradients & 4.93 & 4.90 & 4.88 & 4.88& 4.88\cr
No Noise & 4.26 & 4.21 & 4.20 & 4.21& 4.21\cr
Low Noise & 9.28 & 9.25 & 9.25 & 9.25 &9.25\cr
Med Noise & 16.37 & 16.35 & 16.35 &16.35 &16.34\cr
High Noise & 21.76 & 21.74 & 21.74 & 21.72& 21.72\cr
\cmidrule{1-6}
Average & 10.26 & 10.23 & \textbf{10.22} &10.22 &10.22\cr
\bottomrule
\end{tabular}
\caption{Effect of coarse normal number $l$.}
\label{ablation_coarse_number} 
\end{table}

\subsubsection{Density of query vectors.}
We explore the effect of the sampling number $M$ of query vectors on the angle field learned by $f_{\theta}$. We report the performance of different $M=[2.5k, 5k, 7.5k, 10k]$ in Table \ref{ablation_density}. The query vector set that is too sparse (``2.5k'') cannot provide enough sample vectors for the network to learn the angle field, while the sets that are too dense (``7.5k'', ``10k'') make the implicit angle function more complicated and make it difficult for the network to learn the correct angle offsets. We found ``5k'' is a proper trade-off.

\subsubsection{Number of coarse normals.}
In Table \ref{ablation_coarse_number}, we conduct experiments on the PCPNet dataset to explore how the coarse normal number $l$ affects coarse normal refinement at the inference time. We report the performance of different $l=[1, 5, 10, 20, 50]$, where we find that optimizing a single coarse normal leads to the degradation of results, and the best accuracy is achieved for the first time with 10 coarse normals. Optimizing more coarse normals requires longer time and more memory without improving the accuracy.

\section{Conclusion}
In this paper, we proposed NeAF to estimate point normals implicitly. We randomly sample the query vectors in a unit spherical space, estimate their angle offsets to ground truth normals, and output the query vector with the smallest angle offset as the estimated normal. To fully leverage the prior learned by NeAF, we refine the predicted normal vector by minimizing the estimated angle offset for more accurate normal estimation. NeAF achieves state-of-the-art performance on the synthetic dataset PCPNet and exhibits good generalization on real scans in SceneNN and Semantic3D. Furthermore, the promising results in the surface reconstruction task with normals estimated by NeAF justify our effectiveness in real applications.

\section{Acknowledgments}
The corresponding author is Yu-Shen Liu. This work was supported by National Key R\&D Program of China (2022YFC3800600, 2020YFF0304100), the National Natural Science Foundation of China (62272263, 62072268), and in part by Tsinghua-Kuaishou Institute of Future Media Data.

\bibliography{references}
\end{document}